\documentclass[10pt,a4paper]{article}
\usepackage[utf8]{inputenc}
\usepackage{amsmath}
\usepackage{amsfonts}
\usepackage{amssymb}
\usepackage{natbib}
\usepackage{authblk}

\author[1]{Andrew J. Landgraf}
\author[2]{Jeremy Bellay}
\affil[1]{Battelle Health and Analytics, Columbus, OH, \texttt{landgraf@battelle.org}}
\affil[2]{Battelle Cyber Innovations, Columbus, OH, \texttt{bellayj@battelle.org}}

\begin{document}
\title{\texttt{word2vec} Skip-Gram with Negative Sampling is a Weighted Logistic PCA}
\date{}

\maketitle

\begin{abstract}
We show that the skip-gram formulation of \texttt{word2vec} trained with negative sampling is equivalent to a weighted logistic PCA. This connection allows us to better understand the objective, compare it to other word embedding methods, and extend it to higher dimensional models.
\end{abstract}

\section*{Background}

\cite{mikolov2013distributed} introduced the skip-gram formulation for neural word embeddings, wherein one tries to predict the context of a given word. Their negative-sampling algorithm improved the computational feasibility of training the embeddings. Due to their state-of-the-art performance on a number of tasks, there has been much research aimed at better understanding it. \cite{goldberg2014word2vec} showed that skip-gram with negative-sampling algorithm (SGNS) maximizes a different likelihood than the skip-gram formulation poses and further showed how it is implicitly related to pointwise mutual information \citep{levy2014neural}. We show that SGNS is a weighted logistic PCA, which is a special case of exponential family PCA for the binomial likelihood.

\cite{cotterell2017explaining} showed that the skip-gram formulation can be viewed as exponential family PCA with a multinomial likelihood, but they did not make the connection between the negative-sampling algorithm and the binomial likelihood.  \cite{li2015word} showed that SGNS is an explicit matrix factorization related to representation learning, but the matrix factorization objective they found was complicated and they did not find the connection to the binomial distribution or exponential family PCA. 

\section*{Weighted Logistic PCA}

Exponential family principal component analysis is an extension of principal component analysis (PCA) to data coming from exponential family distributions. Letting $\textbf{Y} = [y_{ij}]$ be a data matrix, it assumes that $y_{ij}, i = 1, \ldots, n, j = 1, \ldots, d$, are generated from an exponential family distribution with corresponding natural parameters $\theta_{ij}$. Exponential family PCA decomposes $\boldsymbol{\Theta} = [\theta_{ij}] = \textbf{A}\textbf{B}^T$, where $\textbf{A} \in \mathbb{R}^{n \times f}$, $\textbf{B} \in \mathbb{R}^{d \times f}$, and $f < \min(n, d)$. This implies that $\theta_{ij} = \textbf{a}_i^T \textbf{b}_j$, where $\textbf{a}_i \in \mathbb{R}^f$ is the $i$th row of $\textbf{A}$ and $\textbf{b}_j \in \mathbb{R}^f$ is the $j$th row of $\textbf{B}$. 

When the exponential family distribution is Gaussian, this reduces to standard PCA. When it is Bernoulli ($y_{ij} \in \{0, 1\}, Pr(y_{ij} = 1) = p_{ij}$), this is typically called logistic PCA and log likelihood being maximized is
\[
\sum_{i,j} y_{ij} \theta_{ij} - \log( 1 + \exp(\theta_{ij})), 
\]
where $\theta_{ij} = \log \left( \frac{p_{ij}}{1 - p_{ij}} \right)$ is the log odds and is approximated by the lower dimensional $\textbf{a}_i^T \textbf{b}_j$.

Just as in logistic regression, when there is more than one independent and identically distributed trial for a given $(i, j)$ combination, the distribution becomes binomial. If there are $y_{ij}$ successes out of $n_{ij}$ opportunities, then the log likelihood is
\[
\sum_{i,j} n_{ij} \left( \hat{p}_{ij} \theta_{ij} - \log( 1 + \exp(\theta_{ij})) \right), 
\]
where $\hat{p}_{ij} = \frac{y_{ij}}{n_{ij}}$ is the proportion of successes. This can be viewed as a \textbf{weighted logistic PCA} with responses $\hat{p}_{ij}$ and weights $n_{ij}$.

\section*{Skip-Gram with Negative Sampling}

SGNS compares the observed word-context pairs with randomly-generated non-observed pairs and maximizes the probability of the actual word-context pairs, while minimizing the probability of the negative pairs.

%
%


Let $n_{w, c}$ be the number of time word $w$ is in the context of word $c$, $n_w$ and $n_c$ be the number of times word $w$ and context $c$ appears, $|D|$ be the number of word-context pairs in the corpus, $P_D(w) = \frac{n_w}{|D|}$, $P_D(c) = \frac{n_c}{|D|}$\footnote{In \cite{mikolov2013distributed} define $P_D(c) \propto n_c^{0.75}$, but without loss of generality, we use the simpler definition in this paper.}, and $P_D(w, c) = \frac{n_{w, c}}{|D|}$ be the distributions of the words, contexts, and word-context pairs, respectively, and $k$ be the number of negative samples. 

Letting $\sigma(x) = \frac{1}{1 + e^{-x}}$, \cite{levy2014neural} showed that SGNS maximizes
\[
\sum_w \sum_c n_{w, c} \left( \log \sigma(\textbf{v}_w^T \textbf{v}_c) + k E_{c^\prime \sim P_D}[\log \sigma(- \textbf{v}_w^T \textbf{v}_{c^\prime})] \right),
\]
where $\textbf{v}_w$ and $\textbf{v}_c$ are the $f$-dimensional vectors for word $w$ and context $c$, respectively.

The SGNS objective can be rewritten
\begin{eqnarray*}
\ell & = & \sum_w \sum_c n_{w, c} \left( \log \sigma(\textbf{v}_w^T \textbf{v}_c) + k E_{c^\prime \sim P_D}[\log \sigma(- \textbf{v}_w^T \textbf{v}_{c^\prime})] \right) \\
& = & \sum_w  \left\{ \left[ \sum_c n_{w, c} \log \sigma(\textbf{v}_w^T \textbf{v}_c) \right] +\left[ \sum_c n_{w, c} k E_{c^\prime \sim P_D}[\log \sigma(- \textbf{v}_w^T \textbf{v}_{c^\prime})]  \right] \right\} \\
& = & \sum_w \left\{  \left[ \sum_c n_{w, c} \log \sigma(\textbf{v}_w^T \textbf{v}_c) \right] + \left[ k n_{w} E_{c^\prime \sim P_D}[\log \sigma(- \textbf{v}_w^T \textbf{v}_{c^\prime})] \right] \right\} \\
& = & \sum_w \left\{  \left[ \sum_c n_{w, c} \log \sigma(\textbf{v}_w^T \textbf{v}_c) \right] + \left[ k n_{w} \sum_{c^\prime} P_D(c^\prime) \log \sigma(- \textbf{v}_w^T \textbf{v}_{c^\prime}) \right] \right\} \\
& = & \sum_w \sum_c \left\{ n_{w, c} \log \left( \frac{\exp(\textbf{v}_w^T \textbf{v}_c)}{1 + \exp(\textbf{v}_w^T \textbf{v}_c)} \right) + k n_{w} P_D(c) \log \left( \frac{1}{1 + \exp(\textbf{v}_w^T \textbf{v}_c)} \right) \right\} \\
& = & \sum_w \sum_c \left\{ n_{w, c} (\textbf{v}_w^T \textbf{v}_c) - (n_{w, c} + k n_{w} P_D(c)) \log \left( 1 + \exp(\textbf{v}_w^T \textbf{v}_c) \right) \right\} \\
& = & \sum_w \sum_c (n_{w, c} + k n_{w} P_D(c)) \left( \frac{n_{w, c}}{n_{w, c} + k n_{w} P_D(c)} (\textbf{v}_w^T \textbf{v}_c) - \log \left( 1 + \exp(\textbf{v}_w^T \textbf{v}_c) \right) \right). \\
\end{eqnarray*}

Define the proportion
\[
x_{w, c} = \frac{n_{w, c}}{n_{w, c} + k n_w P_D(c)} = \frac{P_D(w, c)}{P_D(w, c) + k P_D(w) P_D(c)}.
\]
Then SGNS maximizes
\[
\sum_w \sum_c (n_{w, c} + k n_w P_D(c)) \left( x_{w, c} (\textbf{v}_w^T \textbf{v}_c) - \log(1 + \exp(\textbf{v}_w^T \textbf{v}_c)) \right),
\]
which is logistic PCA with responses $x_{w, c}$ and weights $(n_{w, c} + k n_w P_D(c))$.

Multiplying by the constant $1 / |D|$, the objective becomes
\[
\sum_w \sum_c (P_D(w, c) + k P_D(w) P_D(c)) \left( x_{w, c} (\textbf{v}_w^T \textbf{v}_c) - \log(1 + \exp(\textbf{v}_w^T \textbf{v}_c)) \right),
\]
which gives the weights a slightly easier interpretation.

\section*{Implications}

\paragraph*{Interpretation}

Interpreting the objective, weights will be larger for word-context pairs with higher number of occurrences, as well as for word and contexts with higher numbers of marginal occurrences. The response $x_{w, c}$ is 0 for all non-observed pairs and will be closer to 1 if the number of word-context pair occurrences is large compared to the marginal word and context occurrences. The number of negative samples per word, $k$, has the effect of regularizing the proportions down from 1. Larger $k$'s will also diminish the effect of the word-context pairs in the weights.

\paragraph*{Comparison to Other Results}

We can easily derive the main result from \cite{levy2014neural}, the implicit factorization of the pointwise mutual information (PMI), under this interpretation. For each combination of $w$ and $c$, there are $n_{w,c}$ positive examples and $k n_w P_D(c)$ negative examples. The maximum likelihood estimate of the probability is $x_{w, c}$. The log odds of $x_{w, c}$ is 
\[
\log\left( \frac{n_{w, c} |D|}{n_w n_c} \right) - \log k = PMI(w, c) - \log k,
\] 
which is the same result as in \cite{levy2014neural}.

\paragraph*{Comparison to Other Methods}

Weighted logistic PCA has been used in collaborative filtering of implicit feedback data by Spotify \citep{johnson2014logistic}, where it was referred to as logistic matrix factorization. \cite{johnson2014logistic} was a modification of a previous method which performed matrix factorization with a weighted least squares objective \citep{hu2008collaborative}. \cite{johnson2014logistic} reported that weighted logistic PCA had similar accuracy to \cite{hu2008collaborative}'s weighted least squares method, but could achieve it with a smaller latent dimension.

With that in mind, we can consider an alternative weighted least squares version of SGNS (SGNS-LS), 
\[
\sum_w \sum_c (P_D(w, c) + k P_D(w) P_D(c)) \left( x_{w, c} - \textbf{v}_w^T \textbf{v}_c \right)^2.
\]
Possible advantages include improved computational efficiency and a further comparison with GloVe \citep{pennington2014glove}, which also uses a weighted least squares objective. Ignoring the word and context bias terms, GloVe's objective is
\[
\sum_w \sum_c f(n_{w, c}) \left( \log n_{w, c} - \textbf{v}_w^T \textbf{v}_c \right)^2,
\]
where $f(n_{w, c})$ is a weighting function, which equals 0 when $n_{w, c}$ is 0, effectively removing the non-observed word-context pairs.

Comparing the two objectives, they both have weights increasing as a function of $n_{w,c}$, but SGNS-LS's weights are dependent on the number of marginal occurrences of the words and contexts. Both methods transform the number of word-context occurrences, SGNS-LS converting it to a proportion and GloVe taking the log. We believe the weighting scheme for SGNS-LS has a conceptual advantage over that of GloVe. For example, let $n_{i, j} = n_{k, l} = 1$ with $n_i \gg n_k$ and $n_j \gg n_l$. GloVe treats them both the same, but SGNS-LS will have $x_{i, j} < x_{k, l}$ and will give more weight to $x_{i, j}$ because $n_{i, j}$ being small is much more unlikely due to random chance than $n_{k, l}$ being small.

\paragraph*{Training}

The connection of SGNS to weighted logistic PCA allows us to conceive of other methods to train the word and context vectors. For example, once the sparse word-context matrix has been created, one can either use the MapReduce framework of \cite{johnson2014logistic} or GloVe's approach: sample elements of the matrix and perform stochastic gradient descent with AdaGrad (and similarly for SGNS-LS, with different gradients). GloVe only samples non-zero elements of the matrix, whereas SGNS(-LS) must sample all elements, because the non-occurrence is important for SGNS.

\paragraph*{Extension}

Finally, with this connection to logistic PCA, SGNS can be extended to include other factors in a higher order tensor factorization, analogous to the extension for skip-gram described in \cite{cotterell2017explaining}. Of particular interest is training document vectors along with the word and context vectors. 

\bibliographystyle{chicago}
\bibliography{word2vec}

\end{document}